\documentclass[letterpaper, 10 pt, conference]{ieeeconf}  % Comment this line out if you need a4paper

\IEEEoverridecommandlockouts                              % This command is only needed if 
\usepackage{graphics} % for pdf, bitmapped graphics files
\usepackage{epsfig} % for postscript graphics files
\usepackage{mathptmx} % assumes new font selection scheme installed
\usepackage{times} % assumes new font selection scheme installed
\usepackage{amsmath} % assumes amsmath package installed
\usepackage{amssymb}  % assumes amsmath package installed
\usepackage{afterpage}
\usepackage{graphicx}
\usepackage{subcaption}
\usepackage[backend=biber,style=numeric-comp,sorting=none, block=none,doi=false]{biblatex}
\usepackage{booktabs}
\usepackage{multirow}
\usepackage{hyperref} 
\title{\LARGE \bf
Integrating Diffusion-based Multi-task Learning with Online Reinforcement Learning for Robust Quadruped Robot Control
}

\author{Xinyao Qin$^{1}$, Xiaoteng Ma$^{1}$, Yang Qi$^{1}$, Qihan Liu$^{1}$, Chuanyi Xue$^{1}$, \\Ning Gui$^{1}$, Qinyu Dong$^{1}$, Jun Yang$^{1}$$^{\dagger}$, Bin Liang$^{1}$$^{\dagger}$% <-this % stops a space
\thanks{${\dagger}$Corresponding authors}% <-this % stops a space
\thanks{$^{1}$Department of Automation, Tsinghua University. Email: \{qinxy24\} @mails.tsinghua.edu.cn}%
}
\begin{document}

\maketitle
\thispagestyle{empty}
\pagestyle{empty}

\begin{figure*}[h]
    \centering
    \includegraphics[width=1.0\textwidth]{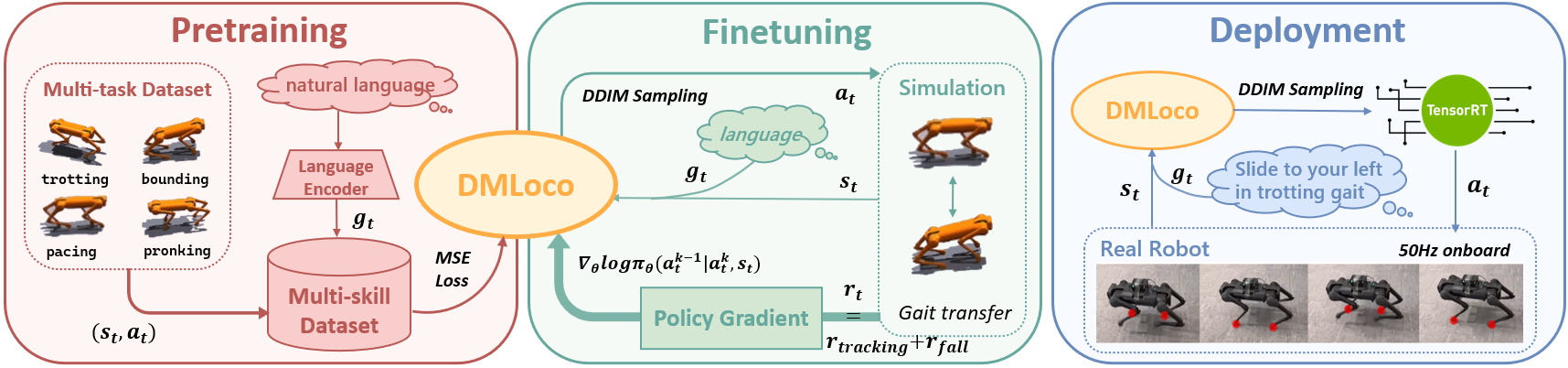}
    \caption{The overall framework of DMLoco. First We perform imitation learning on an offline multi-skill dataset and encode natural language inputs into vectors (Left). Then we preform gait transition with simple rewards in the simulation environment and use the policy gradient algorithm for policy updates (Middle). Finally DMLoco is deployed on real robot with 50Hz control frequency.}
    \label{frame}
    \vspace{-0.5cm}
    \end{figure*}
%%%%%%%%%%%%%%%%%%%%%%%%%%%%%%%%%%%%%%%%%%%%%%%%%%%%%%%%%%%%%%%%%%%%%%%%%%%%%%%%
\begin{abstract}
Recent research has highlighted the powerful capabilities of imitation learning in robotics. Leveraging generative models, particularly diffusion models, these approaches offer notable advantages such as strong multi-task generalization, effective language conditioning, and high sample efficiency. While their application has been successful in manipulation tasks, their use in legged locomotion remains relatively underexplored, mainly due to compounding errors that affect stability and difficulties in task transition under limited data. Online reinforcement learning (RL) has demonstrated promising results in legged robot control in the past years, which can provide valuable insights to address these challenges.
In this work, we propose DMLoco, a diffusion-based framework for quadruped robots that integrates multi-task pretraining with online PPO finetuning to enable language-conditioned control and robust task transitions. Our approach first pretrains the policy on a diverse multi-task dataset using diffusion models, enabling language-guided execution of various skills. Then, the policy is finetuned in simulation to ensure robustness and stable task transition for real-world deployment. By utilizing Denoising Diffusion Implicit Models (DDIM) for efficient sampling and TensorRT for optimized deployment, our policy runs onboard at 50Hz, offering a scalable and efficient solution for adaptive, language-guided locomotion on resource-constrained robotic platforms. The code and video can be found: \href{https://github.com/queenxy/DMLoco}{https://github.com/queenxy/DMLoco}
\end{abstract}

%%%%%%%%%%%%%%%%%%%%%%%%%%%%%%%%%%%%%%%%%%%%%%%%%%%%%%%%%%%%%%%%%%%%%%%%%%%%%%%%
\section{Introduction}
Recent advances in generative modeling, particularly diffusion models, have demonstrated remarkable capabilities in capturing complex, high-dimensional, and multi-modal data distributions. Their success has been widely documented in domains such as text-to-image generation \cite{ldm, sd3}, video generation \cite{sora}, and speech synthesis \cite{tts}, where they have set new state-of-the-art benchmarks. In the field of robotics, diffusion-based imitation learning has been extensively applied to robotic manipulation tasks from small-scale policies \cite{umi, diffusionpolicy, DP3, dppo} to large Vision-Language-Action models \cite{pi0, rdt, agibot}. These models have achieved significantly superior performance compared to previous methods, especially in language-conditioned control and visual understanding tasks.

Despite these excellent attributes, the application of diffusion models to multi-task learning in legged locomotion is still relatively unexplored. This gap stems from several challenges specific to legged robots. First, the complex and highly nonlinear dynamics of quadruped systems make imitation learning difficult, as accumulated errors during training often result in unstable motion. Second, fast and robust transitions between tasks are required in locomotion tasks, which are challenging due to the scarcity of data capturing intermediate transition phases. Finally, the high computational cost of diffusion models poses a major barrier to real-time deployment, especially on resource-limited legged platforms that depend on high-frequency control.

For these issues, online reinforcement learning (RL) provides a promising solution. Powered by advanced parallel simulation platforms \cite{Genesis, isaaclab, isaacgym, sapien, mujoco}, RL agents continuously trial and error with the environment and improve their policies via reward signals. They have demonstrated impressive sim-to-real transfer capabilities, enabling legged robots to achieve stable locomotion across diverse terrains \cite{terrains, eth} and transition smoothly between multiple gaits \cite{wtw, amp}. However, such approaches often depend on carefully designed and complex reward functions, while also suffering from relatively low sample efficiency. Furthermore, they tend to be less compatible with language-conditioned and vision-based tasks, which become increasingly important in robot control.

In this work, we propose \textbf{DMLoco}, a novel framework that integrates \textbf{D}iffusion-based \textbf{M}ulti-task pretraining with online Proximal Policy Optimization (PPO) finetuning for quadruped \textbf{Loco}motion. Our approach first learns a language-conditioned policy from diverse offline expert datasets utilizing the strong representational capacity of diffusion models. Then the policy is finetuned by online PPO to enhance robustness and enable smooth gait transitions without access to explicit transition data. By leveraging Denoising Diffusion Implicit Models (DDIM) for efficient sampling and TensorRT for optimized onboard deployment, our policy can finally achieve real-time inference at 50 Hz, making it readily applicable for real-world robotic platforms.

Our main contributions are as follows:

1) A novel quadruped multi-task learning framework that integrates diffusion models with PPO fine-tuning, allowing the system to effectively learn diverse skills while maintaining strong robustness.

2) Stable task transitions in both simulation and real robot, despite the absence of explicit gait transition data in the training set.

3) A language-conditioned control interface that facilitates improved human-robot interaction and shows promising generalization to unseen natural language commands.

4) Successfully deploy our approach on a physical quadruped robot, achieving real-time onboard inference at 50 Hz, thereby validating its practicality on resource-constrained hardware.

\section{Related works}
\subsection{Multi-skill learning for Locomotion}
Multi-skill learning remains a central topic in legged robotics. A simple idea is training policies individually and subsequently distilling them into a compact model \cite{experts,rldg}; however, the inherent coordination difficulty of distillation often leads to unstable performance as the number of skills increases. Another strategy utilizes contact-related rewards such as probabilistic periodic costs based on contact forces and velocities to induce different gaits \cite{wtw,r2,rewardshaping}. However, they are difficult to extend to non-gait skills and necessitate carefully crafted reward functions. The challenge of balancing velocity tracking with complex gait behaviors can hinder adaptability to varying environments and speeds. Adversarial Motion Priors (AMP) \cite{amp} offer flexible, data-driven guidance for achieving natural motion styles by utilizing style-reward signals. While AMP does not rely on explicit kinematic constraints, it can suffer from mode collapse when trained on diverse motion clips due to its GAN-style networks. And it requires latent variables for skill guidance \cite{cassi,latent, skill} which entail more complex training to conduct explicit control instructions. In contrast, DMLoco leverages the multi-modal capacity of diffusion models to ensure coverage for each skill and achieves comparable robustness to established RL methods through subsequent finetuning.

\subsection{Diffusion Models in Robots}
Recent advances have led to the growing applications of diffusion models in robotics, incorporating techniques such as imitation learning \cite{diffuseloco,diffusionpolicy,DP3,umi,disco,render,consistency,3d}, reward learning \cite{diffusionreward,extracting}, data generation, online and offline reinforcement learning \cite{dppo}, and motion generation \cite{disney}. Among these approaches, representing policies as conditional diffusion models has achieved significant success in real-world tasks. However, previous works have primarily focused on high-level planning for manipulation tasks, which are characterized by low-dimensional action spaces, low replanning frequencies, and inherently more stable dynamics. While \cite{diffuseloco} has explored in locomotion tasks and increased the control frequency to 30Hz, it still remains below the typical control frequencies required for quadruped robots and continues to exhibit lower robustness compared to expert policies. In this work, we employ a two-stage training approach to enhance robustness and leverage DDIM to achieve a higher inference frequency of 50 Hz.

\begin{figure*}[h]
    \centering
     \includegraphics[width=1.0\textwidth]{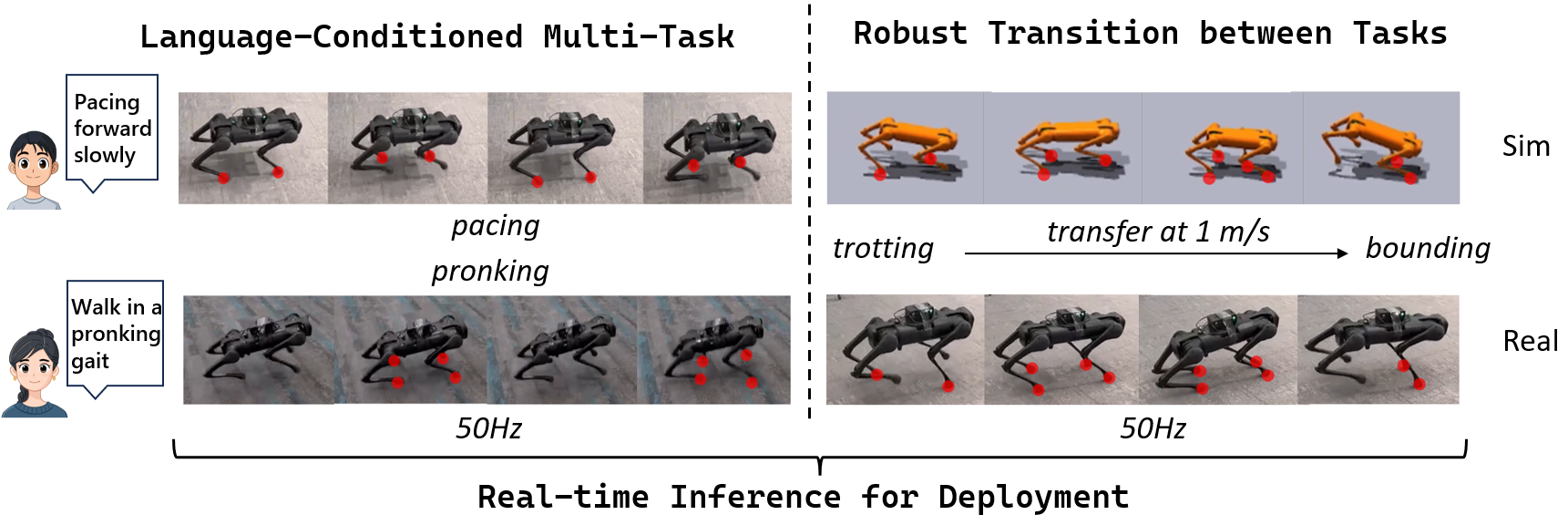}
    \caption{Snapshots of DMLoco showing different gaits, with red points indicating ground contact points. This figure highlights the three core features of DMLoco: Language-Conditioned Multi-task Control, Robust Transition between tasks, and Real-Time Inference onboard.}
    \label{snapshots}
    \vspace{-0.5cm}
    \end{figure*}
    
\section{Preliminaries}
\subsection{Multi-task Imitation learning} \label{3a}
We formulate the problem of learning a multi-task policy $\pi_{\theta}(a|s,g)$ as a goal-conditioned imitation learning problem, where the policy generates the next action $a$ conditioned on the current state $s$ and a task descriptor $g$ (referred to as the goal in the subsequent sections). Here, $g$ can be either a structured instruction comprising velocity commands and one-hot labels representing the gait type or a free-form natural language instruction $l$ embedded into a continuous space via a language encoder $E_l$:
$$
g = \begin{cases} 
[v_x, v_y, \omega, y] & \text{structured instruction}, \\
E_l(l) &  \text{language instruction},
\end{cases}
$$  
where $y$ is the one-hot encoding of the gait type.

Consequently, the policy is trained via supervised learning on a dataset $D = \{\tau_i\}$ consisting of expert state-action trajectories $\tau_i=\{(s_i^0,g_i,a_i^0), ...\}$, with the loss function defined as:
$$
L = \mathbb{E}_{(s, g, a) \sim D} \left[ \|a - \pi_{\theta}(s, g)\|^2 \right]
$$

The language encoder $E_l$ is trained to align the semantic meaning of natural language instructions with the corresponding structured instructions, ensuring that the policy can generalize across unseen languages.
$$
L_{lang} = \mathbb{E}_{(g,l) \sim D} \left[ \|E_l(l) - g_{struc}\|^2 \right]
$$

\subsection{Diffusion Models as Policies} \label{ddpm}
We adopt a UNet-based Denoising Diffusion Probabilistic Model (DDPM) \cite{ddpm} to model the robot's policy. 
The denoising process in DDPMs is typically described as Stochastic Langevin Dynamics, expressed by the following equation:
$$
x_{k-1} = \alpha (x_k - \gamma \epsilon_\theta(x_k, k) + \mathcal{N}(0, \sigma^2 I)),
$$
where $k$ denotes the current time step, $\mathcal{N}(0, \sigma^2 I)$ represents Gaussian noise with mean zero and variance $\sigma^2$, $\epsilon_\theta(x_k, k)$ models the predicted noise correction term by a neural network, and $\alpha$ and $\gamma$ are hyperparameters.

We formulate the policy as a diffusion process conditioned on the robot's current state and goal, expressed as follows:
$$
a_{k-1}^t = \alpha \left(a_k^t - \gamma \epsilon_\theta(S_t, G_t, a_k^t, k)\right) + \mathcal{N}(0, \sigma^2 I)
$$
where $a_k^t$ represents the robot's action at time $t$ and diffusion step $k$ , $S_t$ and $G_t$ denote the state sequence $s_{t-h_s}:s_{t}$ and goal sequence $g_{t-h_s}:g_{t}$ at time $t$, $h_s$ means state horizon which keeps same for $s$ and $g$. Notably, we do not adopt the action chunking technique \cite{act} commonly used in diffusion-based policies. This design choice is motivated by the critical need for agile reactions to environmental changes in legged locomotion, which requires real-time action updates at high frequency based on the robot’s current state. Therefore, our model predicts only a single action at the current timestep. Further discussion can be found in the ablation studies \ref{ablation}.

\subsection{Policy Optimization for Diffusion} 
DDPM can be formulated as a two-layer Markov Decision Process (MDP), where the outer layer corresponds to the environment MDP and the inner layer represents the denoising MDP. This formulation enables policy optimization for diffusion-based policies through policy gradient methods, such as PPO. Let

% The Diffusion MDP uses indices ̄$\bar{t}(t, k) = tK + (K-k-1)$, and performs states, actions and rewards as below,
$$\bar{t}(t, k) = tK + (K-k-1)$$
$$\bar{s}_{\bar{t}(t,k)}=(s_t,a_t^{k+1}),\quad\bar{a}_{\bar{t}(t,k)}=a_t^k$$
% $$\begin{aligned}\quad\bar{R}_{\bar{t}(t,k)}(\bar{s}_{\bar{t}(t,k)},\bar{a}_{\bar{t}(t,k)})=\begin{cases}0&k>0\\R(s_t,a_t^0)&k=0&\end{cases}\end{aligned}$$

The policy and gradient can be written as 
$$\bar{\pi}_\theta(\bar{a}_{\bar{t}}\mid\bar{s}_{\bar{t}})=\mathcal{N}(a_t^k;\mu(a_t^{k+1},\varepsilon_\theta(a_t^{k+1},k+1,s_t)),\sigma_{k+1}^2\mathrm{I})$$
$$\nabla_\theta{J}(\bar{\pi}_\theta)=\mathbb{E}^{\bar{\pi}_\theta}\left[\sum\nabla_\theta\log\bar{\pi}_\theta(\bar{a}_{\bar{t}}\mid\bar{s}_{\bar{t}})\bar{r}(\bar{s}_{\bar{t}},\bar{a}_{\bar{t}})\right]$$
Thus, we can utilize policy gradient algorithms in general reinforcement learning to optimize the diffusion-based policy by directly maximizing the expected cumulative reward \cite{dppo}. This approach enables the policy to be finetuned after learning from an offline dataset, enhancing its adaptability and robustness in real-world scenarios.

\section{Methods}
In this section, we present a detailed introduction to the framework and technical components of DMLoco. DMLoco begins by constructing a multi-task expert dataset, which is then utilized for imitation learning to extract a multi-task policy. This policy is subsequently finetuned using the policy gradient method to enhance robustness. Additionally, DMLoco integrates a natural language encoding module enabling the understanding of language instructions. During deployment, DMLoco uses DDIM for faster sampling and TensorRT for efficient onboard deployment. The overall framework is illustrated in Fig. \ref{frame}, and a detailed explanation of these modules is provided below.
\subsubsection{Dataset}
In this study, we employ single-task policies trained via PPO to collect expert demonstration data. It is important to note that our approach is not restricted to a specific source of expert policies -- these policies can be derived from various learning algorithms or even traditional control strategies. For the quadruped robot, we select four distinct gaits as separate tasks: trotting, bounding, pacing, and pronking. The robot’s observation space includes angular velocity, gravity vector, joint positions and velocities, and the previous action, while the action space corresponds to the desired proportional-derivative (PD) targets for each joint. The task condition is provided in both structured and unstructured formats as Section \ref{3a}. To enhance the policy’s robustness, we incorporate domain randomization during data collection by varying key parameters such as PD gains, system mass, and ground friction coefficients.

\subsubsection{Pretraining}
\begin{figure}[t]
    \centering
     \includegraphics[width=0.5\textwidth]{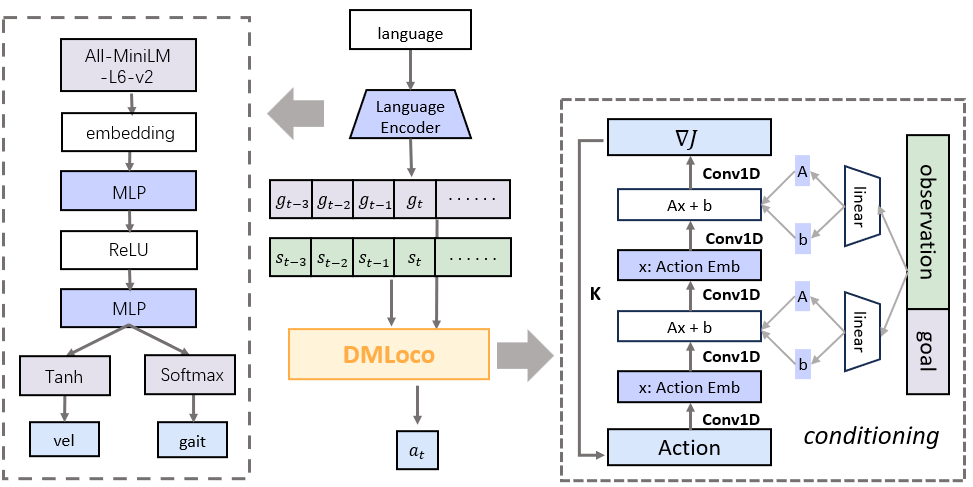}
    \caption{Network architecture of DMLoco. (Left) Natural language inputs are processed by a pretrained language model, All-MiniLM-L6-v2, and then aligned with velocity commands and gait labels through a two-layer MLP network. (Right) At time step $t$, DMLoco takes an observation sequence and a goal sequence as conditional inputs. After performing $K$ steps of denoising, it generates an action sequence and executes the first action.}
    \label{network}
    \vspace{-0.5cm}
    \end{figure}

In the following, we adopt a diffusion model as the policy to perform behavior cloning on offline datasets. Unlike the standard formulation of diffusion policies \cite{diffusionpolicy}, we retain sequences of states and goals as inputs, but predict actions at individual time steps. During training, we randomly sample a diffusion step $k$ from $\{1, ..., K\}$ and inject Gaussian noise $\epsilon_k$ into the ground truth action. A U-Net-based denoising model then takes the noisy action, the state trajectory, the goal trajectory, and the diffusion step $k$ as inputs and predicts the noise component $\epsilon_{\theta}$, as illustrated in Fig. \ref{network}. The model is trained to minimize the mean squared error (MSE) between the predicted noise $\epsilon_{\theta}$ and the ground truth noise $\epsilon_k$. Through this process, the policy learns to generate actions conditioned on the robot's state and goal sequences from the dataset. Through this process, the policy learns to generate actions conditioned on the robot’s state and goal sequences from the dataset. This design enables the policy to adapt to rapidly changing robot states, ensuring that the executed actions remain responsive to the latest observations and thus contribute to stable execution.

\subsubsection{Finetuning} 
To further refine the policy and enhance its stability on the physical robot, we adopt an online reinforcement learning approach. While the pretrained policy demonstrates satisfactory performance in following velocity commands for walking on flat terrain, it struggles with gait transitions due to the lack of relevant data in the training dataset. This limitation often leads to instability, particularly during high-speed gait transitions, resulting in the robot falling. During finetuning, trajectories are collected through interactions with the simulated environment, and the policy is iteratively updated using the PPO algorithm based on computed advantage functions. Compared to pre-training phase, we employ a relatively high diffusion noise scheduling to encourage exploration. 
In contrast to prior methods for quadrupedal robots that rely on complex reward functions composed of dozens of terms \cite{wtw}, we simplify the reward function during finetuning to include only two key components: speed tracking error and fall detection. This streamlined approach minimizes the need for extensive reward engineering while ensuring that the proposed method remains broadly applicable and computationally efficient.

\subsubsection{Language Embedding}\label{language}
We train our language encoders on top of the pretrained language embeddings from all-MiniLM-L6-v2 \cite{minilmv2}, a lightweight yet powerful language model. Specifically, natural language inputs are first processed through the pretrained language model to generate 384-dimensional representation vectors. These vectors are then passed through a two-layer multi-layer perceptron (MLP) for dimensionality reduction, producing compact and task-aligned embeddings. The resulting embeddings are concatenated with the robot's observation vectors and fed into the diffusion model as part of the conditional input. To ensure robust language generalization, we generate a diverse set of 1,000 free-form natural language instructions in English using GPT-4. The dataset is then split into training, validation, and test sets in an 8:1:1 ratio to facilitate effective model evaluation and prevent overfitting. 

\subsubsection{Real-Time Inference}
To achieve efficient inference, we leverage DDIM sampling in both finetuing and deployment. In contrast to DDPM, DDIM modifies the reverse diffusion process by adopting a deterministic and non-random sampling approach, enabling the generation of high-quality samples with significantly fewer sampling steps. Unlike \cite{diffuseloco}, which relies on DDPM for sampling, we demonstrate that DDIM-based sampling not only enhances inference speed but also plays a critical role in improving the robustness of robotic motion by mitigating delayed input issues. Furthermore, to address the constraints of limited onboard computational resources, we employ TensorRT for accelerated deployment, ensuring real-time performance on embedded hardware platforms.

\begin{table*}[h]
\setlength{\tabcolsep}{10pt}
\caption{\textbf{Velocity-tracking Performance across Different Baselines and DMLoco}}
\centering
\begin{tabular}{llcccccc}
\toprule
Gait&Metrics&BeT&DiffuseLoco&WTW&Cassi&DMLoco&DMLoco-lang \\
\midrule 
\multirow{2}{*}{Trotting}&Success Rate($\%$)&94&100&100&100&\textbf{100}&100 \\
&Tracking error($m^2/s^2$)&0.23&0.15&0.14&0.14&\textbf{0.10}&0.17\\
\midrule 
\multirow{2}{*}{Bounding}&Success Rate($\%$)&91&98&100&100&\textbf{100}&100 \\
&Tracking error($m^2/s^2$)&0.22&0.16&0.11&0.13&\textbf{0.10}&0.14 \\
\midrule 
\multirow{2}{*}{Pacing}&Success Rate($\%$)&88&99&100&98&\textbf{100}&100 \\
&Tracking error($m^2/s^2$)&0.25&0.16&0.16&0.18&\textbf{0.13}&0.13 \\
\midrule 
\multirow{2}{*}{Pronking}&Success Rate($\%$)&86&98&100&99&\textbf{100}&100 \\
&Tracking error($m^2/s^2$)&0.42&0.36&0.27&0.38&\textbf{0.31}&0.33 \\
\bottomrule
\specialrule{0em}{2pt}{2pt}
\multirow{2}{*}{Total}&Success Rate($\%$)&89.75&98.75&100.00&99.25&\textbf{100.00}&100.00 \\
&Tracking error($m^2/s^2$)&0.28&0.21&0.17&0.21&\textbf{0.14}&0.19 \\
\bottomrule
\end{tabular}
\label{table1}
\vspace{-0.3cm}
\end{table*}

\section{Experiments}
In this section, we evaluate the performance of DMLoco and conduct ablation experiments to analyze its key components. To make a comprehensive comparison, we selected two kinds of baselines: imitation learning and multi-skill RL. Our experiments are conducted in both simulation and real-world environments. Simulations are performed in Isaac Gym on a Nvidia GeForce RTX 4080 laptop with Intel i9 14900kf, while physical experiments are carried out on a Unitree Aliengo robot equipped with an NVIDIA Jetson Orin NX computing platform. The setup also includes a microphone for speech recognition, utilizing OpenAI's Whisper model for speech-to-text conversion. All the language-conditioned experiments are conducted on the test dataset generated in \ref{language} which is unseen during training.

\subsection{Tasks and Baselines}\label{tasks}
To evaluate the language-conditioned control capability and robustness of DMLoco, we conduct two different tasks below. The first one is velocity-tracking tasks across four distinct gaits (trotting, bounding, pacing, and pronking) at varying velocity commands. In the velocity tracking task, the target v is randomly selected from the following ranges: \(v_x \in [-1.0, 1.0] \, \text{m/s}\), \(v_y \in [-0.6, 0.6] \, \text{m/s}\) , and \(\omega_z \in [-0.5, 0.5] \, \text{rad/s}\). The experiment is conducted in simulation with repeat of 100 times for each gait. 

We also test the robustness of DMLoco in the gait transition task. This task is tested under three different speeds: 0.1 m/s, 0.5 m/s, 1.0 m/s on the real robot. We test four types of gait transitions: trotting to bounding, bounding to pacing, pacing to pronking, and pronking to trotting. Each transition type is repeated 5 times, resulting in 20 trials in total. 

To provide a comprehensive comparison, we select four state-of-the-art baselines: BeT \cite{bet}, DiffuseLoco \cite{diffuseloco}, WTW \cite{wtw} and Cassi \cite{cassi}. These baselines include imitation learning methods and multi-skill RL methods, which utilize different kinds of models and styles of rewards. 

Details of all the algorithms are described below. 
\begin{itemize}
    \item \textbf{BeT}: It employs an autoregressive transformer to learn a policy from an offline multi-modal dataset. It operates by predicting discrete action bins along with their corresponding continuous offsets.
    \item \textbf{DiffuseLoco}: A diffusion-based multi-task policy learning from offline dataset which is similar to our pretraining stage but lacks of online finetuning.
    \item \textbf{WTW}: It designs complex reward functions for multi-gait locomotion by calculating periodic contact forces and velocities. The policy is subsequently optimized using PPO.
    \item \textbf{Cassi}: A framework that integrates unsupervised skill discovery into an AMP-style training procedure, enabling explicit skill selection via a latent skill variable.
    \item \textbf{DMLoco (ours)}: We propose this two-stage method that combines diffusion-based multi-task pretraining on an offline dataset with online fine-tuning using a simple reward function in simulation.
    \item \textbf{DMLoco-lang (ours)}: A variant of DMLoco trained with language instructions rather than structured commands, demonstrating the potential for more flexible control over non-diffusion approaches.
\end{itemize}

\begin{figure}[h]
    \centering
     \includegraphics[width=0.38\textwidth]{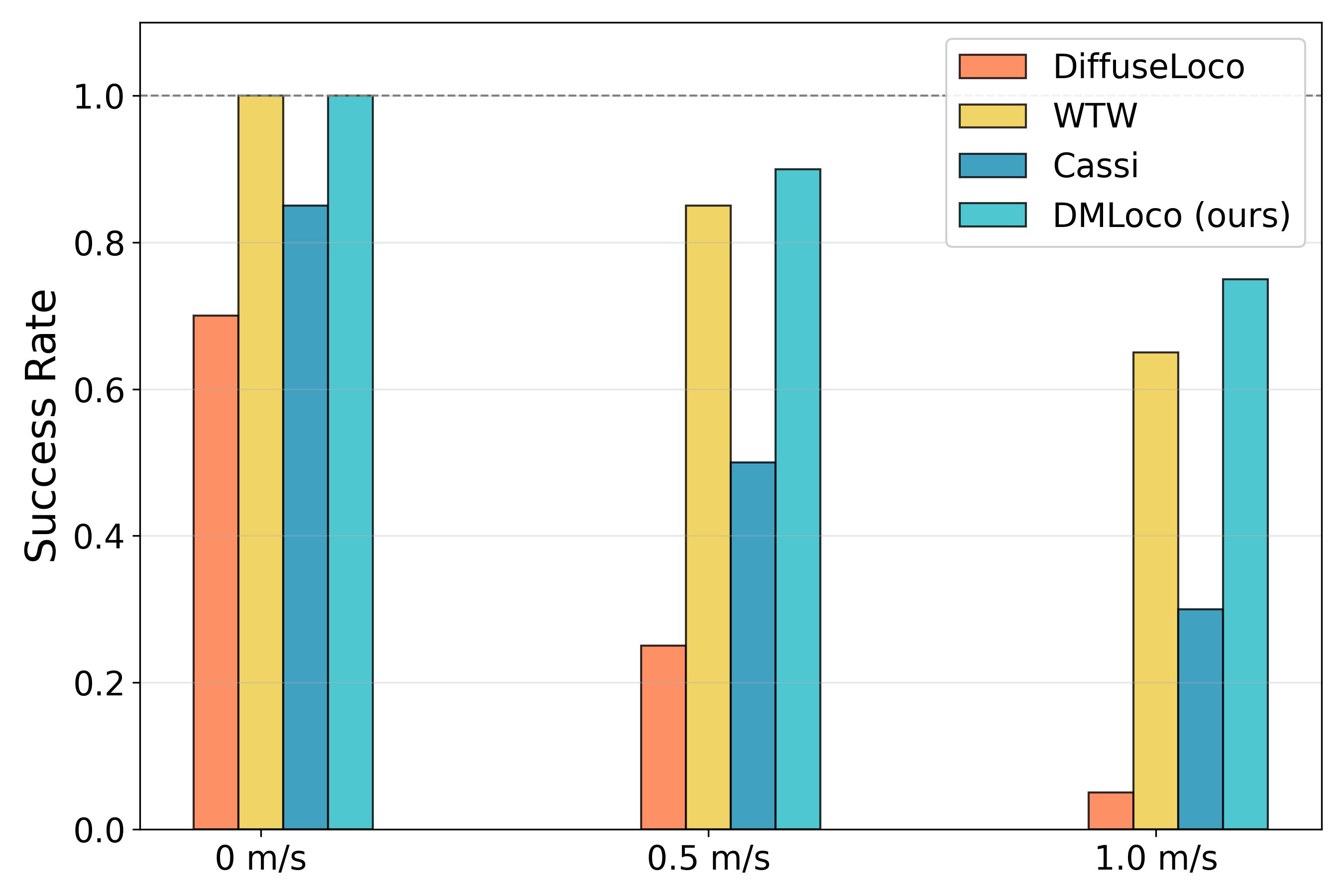}
    \caption{Gait Transition Success Rate over Baselines}
    \label{gait}
    \vspace{-0.7cm}
    \end{figure}

\subsection{DMLoco versus baselines}
For the velocity-tracking experiment, performance is evaluated using two key metrics: Success Rate (higher is better), defined as the percentage of trials in which the robot correctly follows the instruction and remains upright throughout the entire 10-second period, and Tracking Error (lower is better), measured as the mean squared error (MSE) between the actual velocity and the target velocity. As summarized in Table \ref{table1}, DMLoco achieves the highest success rate and the lowest tracking error among all baseline methods. Compared to pure offline imitation learning, DMLoco enhances execution stability through its online fine-tuning process. It also improves velocity tracking performance via a simplified reward function that focuses exclusively on tracking accuracy and fall prevention. In contrast, WTW relies on a complex reward structure that requires the robot to simultaneously optimize multiple objectives, while Cassi necessitates additional training of a skill discriminator. These factors divert attention from velocity tracking and thus slightly hinder performance. We also observe that compared to DMLoco trained on structured instructions, DMLoco-lang achieves a comparable success rate with only a marginal increase in velocity tracking error. This result highlights the strong representational capacity and generalization ability of the diffusion-based approach.

The gait transition success rate is illustrated in Figure \ref{gait}. Due to the poor performance of BeT and the language recognition latency of DMLoco-lang during task transitions, we only report the results for DiffuseLoco, WTW, Cassi, and DMLoco (ours). Experimental results show that DMLoco achieves the highest gait transition success rate among all evaluated methods. Owing to the lack of environmental interaction, DiffuseLoco exhibits a lower transition success rate compared to RL-based methods, and its performance declines rapidly as speed increases—almost failing completely at 1.0 m/s. Cassi learns stable locomotion implicitly by guiding the robot to imitate reference motions, rather than incorporating explicit fall-prevention mechanisms. As a result, it suffers from a lower success rate at higher velocities. Although WTW, like DMLoco, explicitly penalizes falls through reward design, its complex reward structure, which requires balancing multiple objectives, slightly hinders its final performance, as discussed earlier.

% \begin{table}[h]
% \setlength{\tabcolsep}{12pt}
% \caption{\textbf{Success Rate of Gait Transition}}
% \centering
% \begin{tabular}{llcc}
% \toprule
% Transfer&\multirow{2}{*}{Env} &\multirow{2}{*}{Pretrain}&\multirow{2}{*}{Finetune}\\
% Speed &&&\\
% \midrule 
% \multirow{2}{*}{low}&sim&85/100&\textbf{100/100} \\
% &real&11/20&\textbf{20/20} \\
% \midrule 
% \multirow{2}{*}{medium}&sim&76/100&\textbf{99/100} \\
% &real&4/20&\textbf{18/20} \\
% \midrule 
% \multirow{2}{*}{high}&sim&63/100&\textbf{91/100} \\
% &real&2/20&\textbf{15/20} \\
% \bottomrule%第四道横线
% \end{tabular}
% \label{gait}
% \vspace{-0.5cm}
% \end{table}

\subsection{Ablation Studies}\label{ablation}
To further validate the effectiveness of various design choices in the DMLoco framework, we conduct some ablation experiments. These experiments aimed to explore the impact of different components on model performance and provide guidance for future improvements.

\begin{figure}[h]
    \centering
     \includegraphics[width=0.5\textwidth]{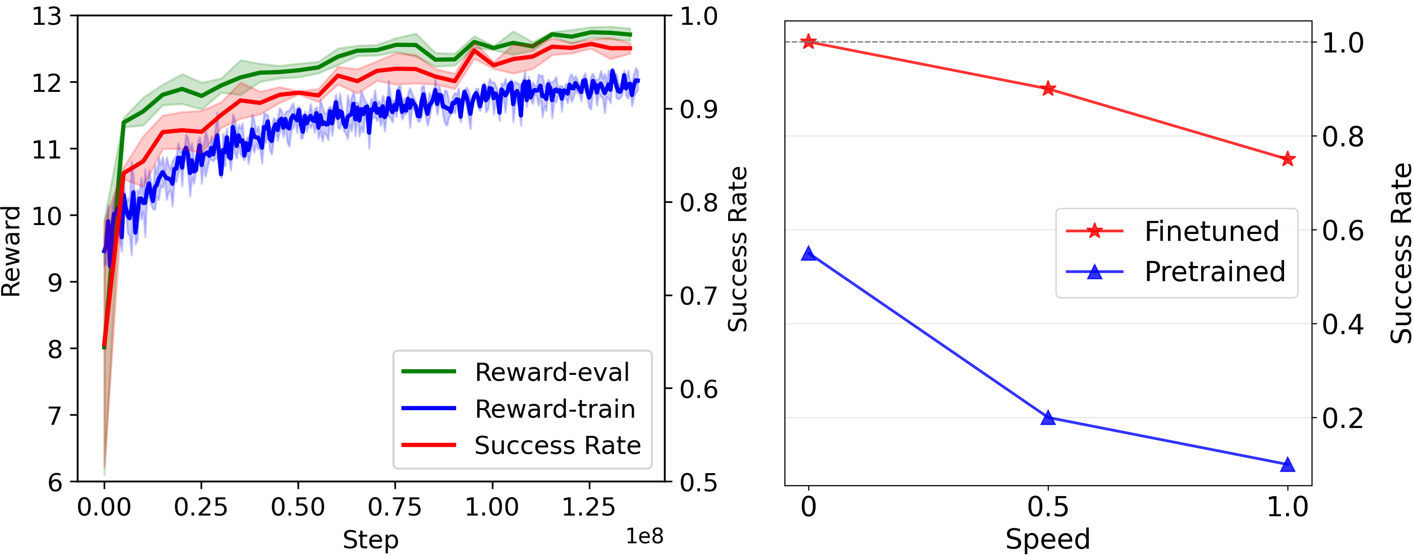}
    \caption{(a) The learning curve of reward and success rate during finetuning across three different seeds. (b) The success rate of gait transition under different speeds for pretrained and finetuned policy.}
    \label{fintune_fig}
    \end{figure}

We first evaluate the improvement brought by fine-tuning. The learning curve of the success rate during simulation training is shown in Figure \ref{fintune_fig}. As illustrated, fine-tuning enhances the policy's performance, increasing the success rate from 0.6 to nearly 1.0. The reward function also exhibits a consistent upward trend throughout the training process. Both the stability and reward increase steadily with low variance across three different random seeds, demonstrating the effectiveness and reproducibility of the finetuning process. 
These improvements are further validated in real-world experiments. We deploy the policies under the same task settings as described in Section \ref{tasks}, and report the result in Figure \ref{fintune_fig}. Compared to the pretrained policy, the finetuned version achieves a 45\% higher success rate at low speed. The performance improvement is even more pronounced at medium and high speeds, with success rates increasing by 70\% and 65\%, respectively. This result clearly demonstrates the significant enhancement in robustness achieved through fine-tuning, especially under highly dynamic conditions.

\begin{figure}[h]
    \centering
     \includegraphics[width=0.48\textwidth]{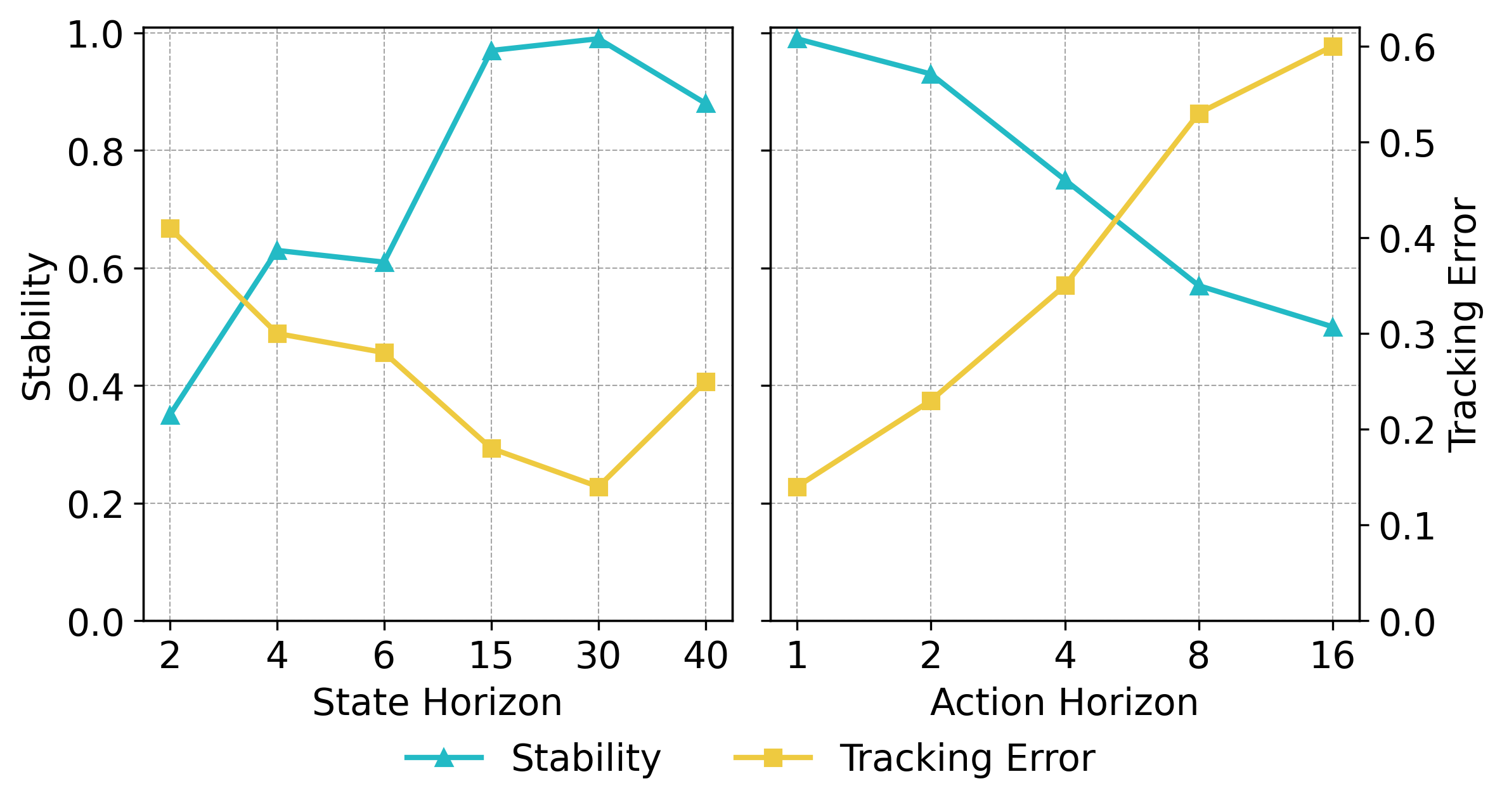}
    \caption{Stability and Tracking Error of DMLoco under different state and action horizons}
    \label{horizon}
    \end{figure}

Next, we examine the influences of state and action horizons on policy performance, which are key parameters in the offline learning phase. The evaluation is conducted using the pretrained policy before fine-tuning in the velocity-tracking task, which sufficiently reveals the impact of these parameters, as our goal is to identify optimal configurations prior to fine-tuning.
The results are illustrated in Fig. \ref{horizon}.
For the state horizon, we fix the action horizon to 1, meaning the policy predicts only the current action, as introduced in \ref{ddpm}. The results indicate that shorter state horizons lead to suboptimal performance, while longer horizons improve decision-making by incorporating richer historical information, which implicitly captures previous motion states. However, excessively long horizons introduce redundancy and reduce model efficiency. For the action horizon, we compare our single-step prediction approach with the action chunk method commonly used in robotic manipulation. The state horizon is fixed to 30, which yielded the best performance in the earlier test. The results show that larger action chunks lead to significant performance degradation, manifested as lower success rates and higher tracking errors. This can be explained by the highly dynamic nature of locomotion, which requires real-time adaptation to the current state. In contrast, action chunks rely on states from several timesteps earlier to generate actions, resulting in delayed responses and unstable performance.

We also compare DDIM and DDPM for policy generation, evaluating their performance in terms of stability and inference speed. The experiments are conducted on a Nvidia GeForce RTX 4080 laptop using Pytorch-fp32, also using the pretrained policy under velocity-tracking task situation. As shown in Table \ref{sampling}, DDIM achieves high stability with only 5 sampling steps, whereas DDPM requires 100 steps to reach a comparable level of stability. The inference speed is measured by the average time required per forward pass. DDIM's inference speed is significantly faster than DDPM's, making it more suitable for real-time control applications. However, reducing DDPM's sampling steps to 5 results in a complete loss of stability (dropping to 0), underscoring DDPM's dependence on high-step sampling for reliable performance. This experiment demonstrates that DDIM's deterministic sampling approach is better suited for quadruped robot control, which requires both fast and stable policy generation.

\begin{table}[h]
\caption{\textbf{Performance of Different Sampling Methods}}
\centering
\begin{tabular}{lcc}
\toprule
Sampling & Stability & Inference Speed (s)\\
\midrule 
DDPM-100/100& 98.75 &0.53\\
DDPM-100/5&0&0.026\\
DDIM-100/5& 99.00 & 0.028\\
\bottomrule%第四道横线
\end{tabular}
\label{sampling}
\vspace{-0.3cm}
\end{table}

\begin{figure}[h]
    \centering
     \includegraphics[width=0.45\textwidth]{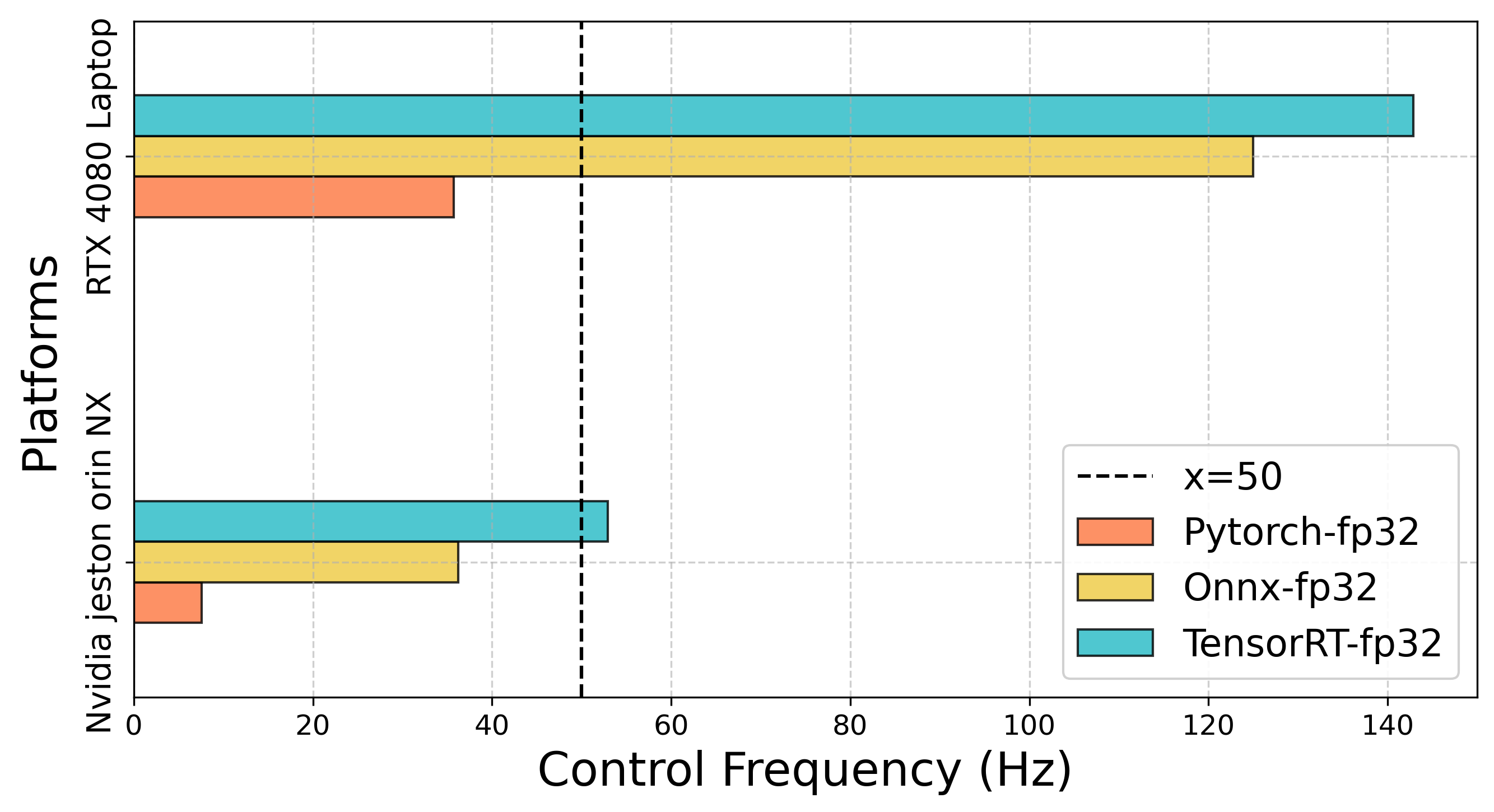}
    \caption{Inference frequency of DMLoco across different platforms and software. 50 Hz is the baseline control frequency required in real time.}
    \label{speed}
    \end{figure}

Finally, we evaluate the inference frequency of DMLoco across different hardware platforms using various software optimizations. As illustrated in Fig. \ref{speed}, TensorRT achieves inference speeds of 142 Hz on a Nvidia GeForce RTX 4080 laptop and 53 Hz on an NVIDIA Jetson, representing improvements of 4× and 7×, respectively, compared to PyTorch. And it also demonstrates superior performance over anothor format ONNX. With TensorRT optimization, DMLoco achieves real-time inference at a control frequency of 50 Hz in both simulation and real-world scenarios, meeting the high-frequency control requirements of quadruped robots. These results underscore the critical role of TensorRT acceleration in enabling efficient and reliable real-time performance during deployment.

\section{Conclusions}
In this work, we presented DMLoco, a diffusion-based multi-task learning framework for quadruped robots that combines generative modeling with reinforcement learning. DMLoco achieves robust, language-conditioned control of diverse locomotion tasks while maintaining real-time performance at 50Hz on resource-constrained hardware. Experiments demonstrated superior stability, language generalization, and robustness, particularly during high-speed gait transitions. While DMLoco shows promising results, its task diversity can be further expanded by incorporating datasets from traditional control methods, and inference efficiency can be improved using other distillation methods. Future work will also explore extending this framework to humanoid robots, broadening its applicability to more complex robotic systems.


%%%%%%%%%%%%%%%%%%%%%%%%%%%%%%%%%%%%%%%%%%%%%%%%%%%%%%%%%%%%%%%%%%%%%%%%%%%%%%%%

%%%%%%%%%%%%%%%%%%%%%%%%%%%%%%%%%%%%%%%%%%%%%%%%%%%%%%%%%%%%%%%%%%%%%%%%%%%%%%%%

%%%%%%%%%%%%%%%%%%%%%%%%%%%%%%%%%%%%%%%%%%%%%%%%%%%%%%%%%%%%%%%%%%%%%%%%%%%%%%%%

\section*{ACKNOWLEDGMENT}
This work was supported by the National Science and Technology Innovation 2030 - Major Project (Grant No. 2022ZD0208804).

\section*{APPENDIX}
The hyperparameters used for training are provided in Table \ref{table2}.

\begin{table}[h]
\setlength{\tabcolsep}{12pt}
\caption{\textbf{Hyperparameters in Training}}
\centering
\begin{tabular}{lcc}
\toprule
Stage & Hyperparameter & Value \\
\midrule 
\multirow{6}{*}{Pretrain} & Batch size & 512 \\
 & Learning rate & 3e-4 \\
 & Denoising steps & 100 \\
 & State horizon & 30\\
 & Diffusion step embed dim & 16 \\
\midrule
\multirow{5}{*}{Finetune} & $\gamma$ & 0.9\\
 & GAE $\lambda$ & 0.95 \\
 & Denoising steps & 5\\
 & Learning rate & 1e-5 \\
 & Sampling denoising std & 0.04 \\
\bottomrule
\end{tabular}
\label{table2}
\vspace{-0.3cm}
\end{table}

\raggedbottom
\printbibliography
\end{document}